\pdfoutput=1

\documentclass[11pt]{article}

\usepackage{acl}

\usepackage{times}
\usepackage{latexsym}

\usepackage[T1]{fontenc}

\usepackage[utf8]{inputenc}

\usepackage{microtype}
\usepackage{algorithm}
\usepackage{algorithmic}
\usepackage{microtype}
\usepackage{times}
\usepackage{latexsym}
\usepackage{booktabs} 
\usepackage{multirow}
\usepackage{url}
\usepackage{amsmath}
\usepackage{bm}
\usepackage{amsfonts}
\usepackage{graphicx}
\usepackage{diagbox}
\usepackage{xspace}
\usepackage{courier}
\usepackage{mathrsfs}
\usepackage{appendix}
\usepackage{float}
\usepackage{framed} 
\usepackage{color}
\usepackage{xspace}
\usepackage{amsfonts}
\usepackage{amssymb}
\usepackage{diagbox}

\usepackage{amsmath}
\usepackage{microtype}
\usepackage{makecell}

\newcommand\base{$_{\small \texttt{BASE}}$\xspace}
\newcommand\bb{$_{\small \texttt{B}}$\xspace}
\newcommand\blarge{$_{\small \texttt{LARGE}}$\xspace}
\newcommand\bl{$_{\small \texttt{L}}$\xspace}
\newcommand\xxlarge{$_{\small \texttt{XXLARGE}}$\xspace}
\newcommand\xxl{$_{\small \texttt{XXL}}$\xspace}
\newcommand\Ourmodel{PL-Marker\xspace}
\usepackage{color}

\title{Packed Levitated Marker for  Entity and Relation Extraction}

 \author{Deming Ye$^{1,2}$, Yankai Lin$^{6}$, Peng Li$^{6,7}$,  Maosong Sun$^{1,2,3,4,5}$\thanks{ \ \ Corresponding author: M. Sun (sms@tsinghua.edu.cn)} \\
$^{1}$Dept. of Comp. Sci. \& Tech., Institute for AI, Tsinghua University, Beijing, China \\
$^{2}$Beijing National Research Center for Information Science and Technology \\
$^{3}$International Innovation Center of Tsinghua University, Shanghai, China \\
$^{4}$Jiangsu Collaborative Innovation Center for Language Ability, Xuzhou, China \\
$^{5}$Institute Guo Qiang, Tsinghua University $^{6}$Pattern Recognition Center, WeChat AI \\
$^{7}$Institute for AI Industry Research (AIR), Tsinghua University\\
\texttt{yedeming001@163.com}
}

\begin{document}

\maketitle

\begin{abstract}

Recent entity and relation extraction works focus on investigating how to obtain a better span representation from the pre-trained encoder. However, a major limitation of existing works is that they ignore the interrelation between spans (pairs). In this work, we propose a novel span representation approach, named Packed Levitated Markers (\Ourmodel),  to consider the interrelation between the spans (pairs) by strategically packing the markers in the encoder. In particular, we propose a neighborhood-oriented packing strategy, which  considers the neighbor spans integrally to better model the entity boundary information. 
Furthermore, for those more complicated span pair classification tasks, we  design a subject-oriented packing strategy, which packs each subject and all its objects to model the interrelation between the same-subject span pairs. 
The experimental results show that, with the enhanced marker feature, our model advances baselines on six NER benchmarks, and obtains a 4.1\%-4.3\% strict relation F1 improvement with higher speed over previous state-of-the-art models on ACE04 and ACE05. 
Our code and models are publicly available at \url{https://github.com/thunlp/PL-Marker}.
 
\end{abstract}

\section{Introduction}

Recently, pre-trained language models (PLMs)~\cite{BERT, RoBERTa} have achieved significant improvements in Named Entity Recognition (NER, \citet{seqner, spanner2}) and Relation Extraction (RE, \citet{DyGIEpp, Typemarker}), two key sub-tasks of information extraction.
Recent works~\cite{AutomatedConcatNER, PURE} regard these two tasks as span classification or span pair classification, and thus focus on extracting better span representations from the PLMs. 


\begin{figure}[!t]
    \centering
    \includegraphics[width=\linewidth]{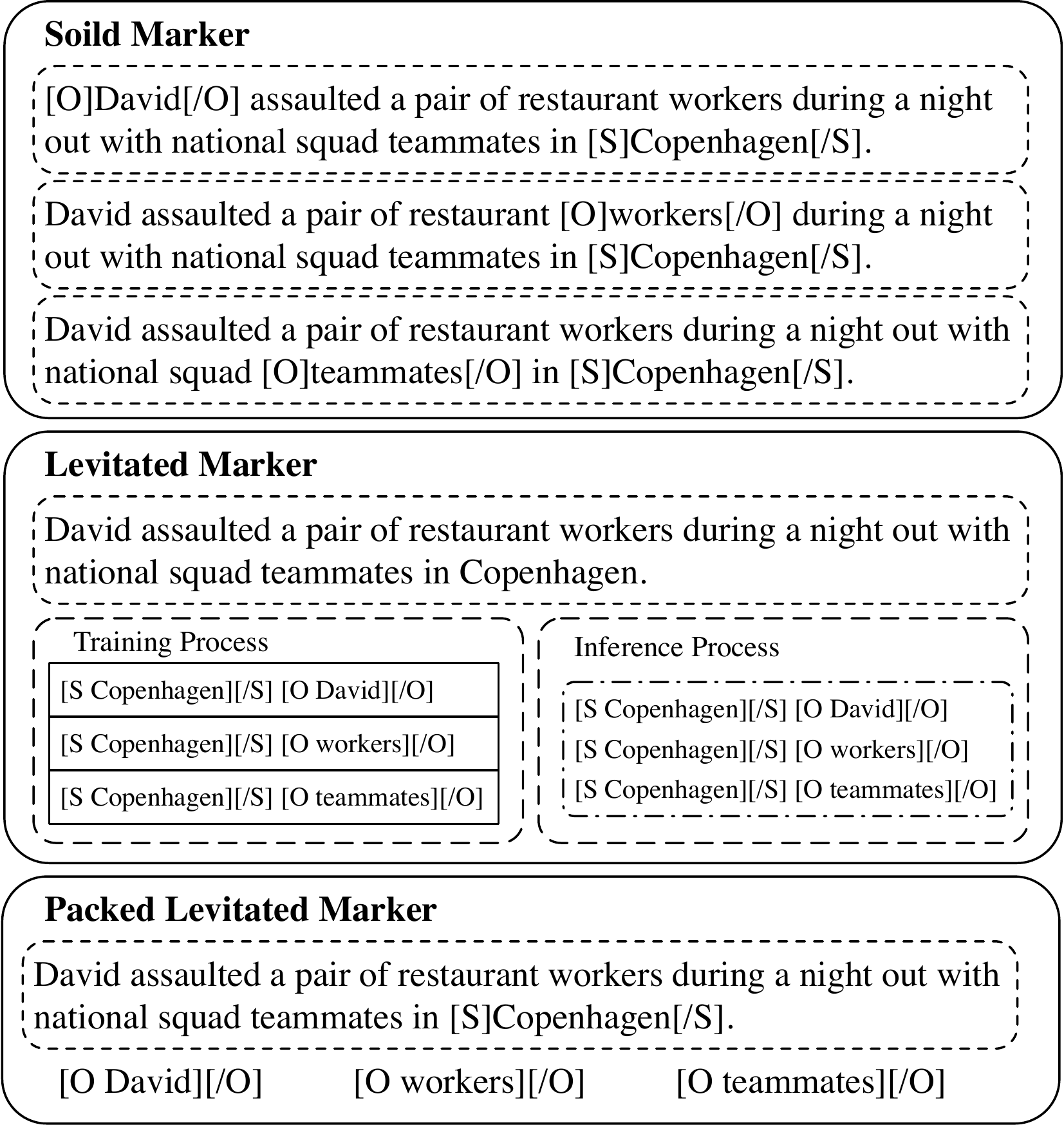}
    \caption{
    An example in the RE task. Solid Marker separately processes three pairs of  spans with different insertions of markers. Levitated Marker  processes the span pairs independently during training and  processes them in batches during inference.  Our proposed Packed Levitated Marker packs three objects for the same subject into an instance to process.
    }
    \label{fig:packing}
\end{figure}

Three span representation extraction methods are widely used: (1) \textbf{T-Concat}~\cite{e2ecoref,  Generalspan} concatenates the  representation of the span's boundary (start and end) tokens to obtain the span representation. It collects information at the token level but ignores the connection between boundary tokens of a span when they pass through the network;  
(2) \textbf{Solid Marker}~\cite{MTB,ChaojunXiao2020DenoisingRE} explicitly insert two solid markers  before and after the span to highlight the span in the input text. And it inserts two pair of markers to locate the subject and object of a span pair. However, the method cannot handle multiple span pairs at the same time because of its weakness in specifying the solid markers of a span pair from more than two pairs of markers in the sequence. 
(3) \textbf{Levitated Marker}~\cite{PURE} first sets a pair of levitated markers to share the same position with the span's boundary tokens and then  ties a pair of markers by a directional attention. To be specific, the markers within a pair are set to be visible to each other in the attention mask matrix, but not to the text token and other pairs of markers. Existing work~\cite{PURE} simply replaces solid markers with levitated markers for an efficient batch computation, but sacrifices the model performance. 
%


As the RE example shown in Figure~\ref{fig:packing}, to correctly identify that  David,  workers and teammates are \emph{located\_in} Copenhagen, it is important to separate out that David \emph{attacked} the  restaurant workers and he had \emph{social} relation with his teammates. However, prior works with markers~\cite{PURE} independently processes the span pairs with different insertions of markers in the training phrase, and thus ignore interrelation between spans (pairs)~\cite{ContextAware, DyGIE, DyGIEpp}. 





In this work, we introduce Packed Levitated Marker (\Ourmodel), to model the interrelation between spans (pairs) by strategically packing levitated markers in the encoding phase. 
A key challenge of packing levitated markers together for span classification tasks is that the increasing number of inserted levitated markers would exacerbate the complexity of PLMs quadratically~\cite{TR-BERT}. Thus, we have to  divide spans into several groups to control the length of each input sequence for a higher speed and feasibility. 
In this case, it is necessary to consider the neighbor spans integrally, which could help the model compare neighbor spans, \emph{e.g.} the span with the same start token, to acquire a more precise entity boundary. Hence, we propose a neighborhood-oriented packing strategy, which packs the spans with the same start token into a training instance as much as possible to better distinguish the entity boundary.  

For the more complicated span pair classification tasks, an ideal packing scheme is to pack all the span pairs together with multiple pairs of levitated markers, to model all the span pairs integrally. 
However, since each pair of levitated markers is already tied by  directional attention,  if we continue to apply directional attention to bind two pairs of markers,  the levitated marker will not be able to identify its partner marker of the same span. 
Hence, we adopt a fusion of solid markers and levitated markers, and use a subject-oriented packing strategy to model the subject with all its related objects integrally.  To be specific,  we emphasize the subject span with solid markers and  pack all its candidate object spans with  levitated markers. Moreover, we apply an object-oriented packing strategy for an intact bidirectional modeling~\cite{corefqa}.

We examine the effect of \Ourmodel on two typical span (pair) classification tasks, NER and end-to-end RE. The experimental results indicate that \Ourmodel with neighborhood-oriented packing scheme performs much better than the model with random packing scheme on NER, which shows the necessity of considering the neighbor spans integrally. And our model  also advances  the T-Concat model on six NER benchmarks, which demonstrates the effectiveness of the feature obtained by span marker. 
Moreover, compared with the previous state-of-the-art RE model, our  model gains  a 4.1\%-4.3\% strict relation F1 improvement with higher speed  on ACE04 and ACE05  and also achieves better performance on SciERC, which shows the importance of considering the interrelation between the subject-oriented span pairs.


\section{Related Work}

In recent years, span representation has attracted great attention from academia, which facilitates  various NLP applications, such as named entity recognition~\cite{instancener}, relation and event extraction~\cite{DyGIE}, coreference  resolution~\cite{e2ecoref}, semantic role labeling~\cite{SRL} and question answering~\cite{qaspan}. Existing methods to enhance span  representation  can be roughly grouped  into three categories: 

\paragraph{Span Pre-training} The span pre-training approaches enhance the span representation for PLMs via span-level pre-training tasks. \citet{ERNIE-baidu,BART, T5} mask and learn to recover  random contiguous spans rather than  random tokens. \citet{spanbert} further learns to store the span information in its boundary tokens for downstream tasks.

\paragraph{Knowledge Infusion} This series of methods focuses on infusing external knowledge into their models.  \citet{ERNIE-zhengyan, KnowBERT, K-adapter} learn to use the external entity embedding from the knowledge graph or the synonym net to acquire knowledge. \citet{MTB,  WKLM,KEPLER,LUKE} conduct specific entity-related pre-training to incorporate knowledge into their models with the help of Wikipeidia anchor texts.

\paragraph{Structural Extension} The structural extension methods add reasoning modules to the existing models, such as biaffine attention~\cite{UniRE}, graph propagation~\cite{DyGIEpp} and memory flow~\cite{TriMF}. With the support of modern pre-training encoders (\emph{e.g.} BERT), the simple model with solid markers could achieve state-of-art results in RE~\cite{Typemarker,PURE}.  
However, it is hard to specify the solid markers of a span pair from more than two pairs of markers in the sequence. 
Hence, previous work~\cite{PURE} has to process span pairs independently, which is time-consuming and ignores the interrelation between the span pairs. In this work, we introduce the neighborhood-oriented and the subject-oriented packing strategies to take advantage of the levitated markers to provide an integral modeling on spans (pairs).

To our best knowledge,  we are the first to apply the levitated markers on the NER. On the RE,  the closest work to ours is the PURE (Approx.)~\cite{PURE},  which  independently encodes each span pair with two pairs of levitated markers in the training phase  and batches multiple pairs of markers to accelerate the inference process. Compared to their work, our model adopts a fusion  subject-oriented packing scheme and thus handle multiple span pairs well in both the training and inference process. We detail the differences between our work and PURE in Section~\ref{sec:compare_pure} and explain why our model performs better.

\section{Method}
In this section, we first introduce the architecture of the levitated marker. Then, we present how we pack the levitated marker to obtain the span representation and  span pair representation.

\begin{figure*}[!t]
    \centering
    \includegraphics[width=0.95\textwidth]{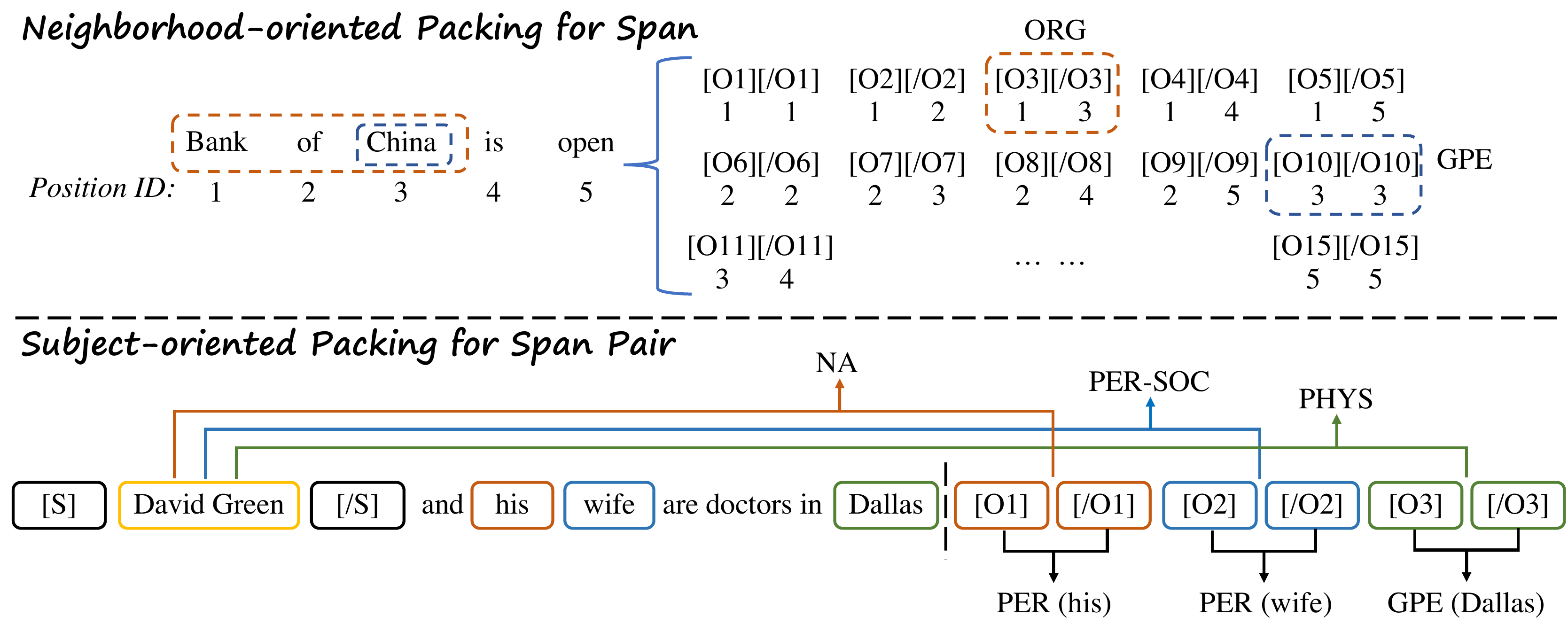}
\caption{An overview of our neighborhood-oriented packing and subject-oriented packing strategies. [S][/S] are solid markers. [O][/O] are levitated markers. With a maximum group size, the neighborhood-oriented packing strategy clusters the neighbor spans, \emph{e.g.} \{(1,1),(1,2),...,(1,5)\}, in the same group.  The subject-oriented packing strategy encloses the subject span, \emph{David Green}, with solid markers, applies levitated markers on its candidate object spans, \emph{his}, \emph{wife} and \emph{Dallas}, and  packs them  into an instance. 
}
    \label{fig:relmodel}
\end{figure*}
\subsection{Background: Levitated Marker}
Levitated marker is used as an approximation of solid markers, which allows models to classify multiple pairs of entities simultaneously to accelerate the inference process~\cite{PURE}.  
A pair of levitated markers, associated with a span, consists of a start token marker and an end token marker. These  two markers share the same position embedding with the start and end tokens of the corresponding span,  while keeping the position id of  original text tokens unchanged. In order to specify multiple pairs of levitated markers in parallel, a directional attention mask matrix is applied. Specifically, each levitated marker is visible to its partner marker within pair in the attention mask matrix, but not to the text tokens and other levitated markers. In the meantime, the levitated markers are able to attend to  the text tokens to aggregate information for their associated spans.

\subsection{Neighborhood-oriented Packing for Span}

Benefiting from the parallelism of levitated markers, we can flexibly pack a series of related spans into a training instance.  In practice, we append multiple associated levitated markers to an input sequence to conduct a comprehensive modeling on each span.  

However, even though the entity length is restricted, some of the span classification tasks still contain a large number of candidate spans. Hence, we have to group the markers into several batches to equip the model with higher speed and feasibility in practice.  To better model the connection between  spans with the same start tokens, we adopt a neighborhood-oriented packing scheme.  As shown in Figure~\ref{fig:relmodel}, we first sort the pairs of levitated markers by taking the position of start marker as the first keyword and the position of end marker as the second keyword. After that, we split them into groups of size up to $K$ and thus gather adjacent spans into the same group. We packs each groups of markers and dispersedly process them in multiple runs. 

Formally, given a sequence of $N$  text tokens, $X = \{x_1, \ldots, x_N \}$ and a maximum span length $L$, we define the candidate spans set as $S(X)=\{(1,1),..,(1,L),...,(N, N-L),..,(N,N))\}$.  
We first divide $S(X)$ into multiple groups up to the size of $K$ in order. For example, we cluster $K$ spans, $\{(1,1),(1,2),...,(\lceil \frac{K}{L} \rceil, K-\lfloor \frac{K-1}{L} \rfloor* L )\}$, into a group $S_1$. 
We associate a pair of levitated markers to each span in $S_1$.  Then, we provide the combined sequence of the text token and  the inserted levitated markers to the PLM (\emph{e.g.} BERT) to obtain the contextualized representations of the start token marker $H^{(s)}=\{h^{(s)}_{i}\}$ and that of the end token marker $H^{(e)}=\{h^{(e)}_{i}\}$. Here,  $h^{(s)}_{a}$ and $h^{(e)}_{b}$ are associated with the span $s_i=(a,b)$, for which we obtain the span representations:
\begin{equation}
     \label{eq:2} \psi(s_i) = [{h}^{(s)}_{a}; {h}^{(e)}_{b}] 
\end{equation}
where $[A; B]$ denotes the concatenation operation on the vector $A$ and $B$.

For instance, we apply the levitated marker to a typical overlapping span classification task, NER, which aims to assign an entity type or a non-entity type to each possible span in a sentence. We obtain the span representation from the PLM via the packed levitated markers and then combine the features of \Ourmodel and T-Concat  to better predict the entity type of the cadidate span.


\subsection{Subject-oriented Packing for Span Pair}

To obtain a span pair representation, a feasible method is to adopt levitated markers to emphasize a series of the subject and object spans simultaneously. Commonly, each pair of levitated markers is tied by the directional attention. But if we continue to apply directional attention to bind two pairs of markers,  the levitated marker will not be able to identify its partner marker of the same span. 
Hence, as shown in Figure~\ref{fig:relmodel}, our span pair model adopts a fusion subject-oriented packing scheme to offer an integral modeling for the same-subject spans. 

Formally, given an input sequence $X$, a subject span, $s_i=(a,b)$ and its candidate object spans $(c_1,d_1), (c_2,d_2), ... (c_m,d_m)$, We insert a pair of solid markers [S] and [/S]  before and after the subject span. 
Then, we apply levitated markers [O] and [/O] to all candidate object spans, and  pack them into an instance. Let $\hat{X}$ denotes this   modified sequence  with  inserted markers:
\begin{align*}
    \hat{X}& = ... \text{[S]}, x_a, ..., x_b, \text{[/S]}, ..., x_{c_1}\cup\text{[O1]}, ..., \\
    &  x_{d_1}\cup\text{[/O1]}, ...  ,x_{c_2}\cup\text{[O2]}, ..., x_{d_2}\cup\text{[/O2]} ...,
\end{align*}
where the tokens jointed by the symbol $\cup$ share the same position embedding. We apply a pre-trained encoder on $\hat{X}$ and finally obtain the span pair representation for $s_i=(a,b)$ and $s_j=(c,d)$:
\begin{equation}
    \phi(s_i, s_j) = [{h}_{{a-1}}; {h}_{{b+1}};{h}^{(s)}_{c};{h}^{(e)}_{d}]
\end{equation}
where $[\,;\,]$ denotes the concatenation operation.  $h_{a-1}$ and $h_{b+1}$ denote the contextualized representation of the inserted solid markers for $s_i$; ${h}^{(s)}_{c}$ and ${h}^{(e)}_{d}$ are the contextualized representation of the inserted levitated markers for $s_j$.

Compared to the method that applies two pairs of solid markers on the subject  and object   respectively~\cite{PURE}, our fusion marker scheme replaces the  solid markers with the levitated markers for the object span, which would impair the emphasis  on the object span to some extent. To provide the supplemental information, we  introduce an inverse relation from the object  to the subject  for a bidirectional prediction~\cite{corefqa}.

For instance, we evaluate our model on a typical span pair classification task, end-to-end RE, which concentrates on identifying whether all span pairs are related and their relation types. Following \citet{PURE}, we first use a NER model to filter candidate entity spans, and then acquire the span pair representation of the filtered entity span pairs to predict the relation between them. Moreover, to build the connection between entity type and relation type, we add an auxiliary loss for predicting the type of object entity~\cite{Typemarker, PTR}.

\subsection{Complexity Analysis}

Dominated by the large feed-forward network, the computation of PLM rises almost linearly with the increase in small sequence length~\cite{Funnel-Transformer,TR-BERT}. Gradually, as the sequence length continues to grow, the computation dilates quadratically due to the Self-Attention module~\cite{Transformer}. 
Obviously, the insertion of levitated markers extends the length of input sequence. 
For the span pair classification tasks, the number of candidate spans is relatively small , thus the increased computation is limited. 
For the span classification tasks,  we group the markers into several batches, which can control the sequence length within the interval in which the complexity increases nearly linearly. For the NER, we enumerate candidate spans in a small-length sentence and then use its context words to expand the sentence to 512 tokens, for which the number of candidate spans in a sentence is usually less than the context length in practice. Hence, with a small number of packing groups, the complexity of \Ourmodel is still near-linearly to the complexity of previous models. 

Moreover, to further alleviate the inference cost, we adopt \Ourmodel as a post-processing module of a two-stage model, in which it is used to identify entities from a small number of candidate entities proposed by a simpler and faster model.  



\section{Experiment}

\subsection{Experimental Setup}

\subsubsection{Dataset}
For the NER task, we conduct experiments on both flat and nested  benchmarks. Firstly, on the flat NER, we adopt CoNLL03~\cite{conll03},  OntoNotes 5.0~\cite{Ontonotes} and Few-NERD~\cite{Fewnerd}. Then, on the nested NER,  we use ACE04~\cite{ACE04}, ACE05~\cite{ACE05} and SciERC~\cite{SciERC}. The three nested NER datasets are also used to evaluate the end-to-end RE. 
We follow \citet{DyGIE} to split ACE04 into 5 folds and split ACE05 into train, development, and test sets. For other datasets, we adopt the official split. 
 Table~\ref{tab:statistics} shows the statistics of each dataset. 



\subsubsection{Evaluation Metrics}

For NER task, we follow a span-level evaluation setting, where the entity boundary and entity type are required to correctly predicted.   For the end-to-end RE, we report two evaluation metrics: (1) \textbf{Boundaries evaluation} (Rel) requires the model to correctly predict the boundaries of the subject entity and the object entity, and the entity relation; (2) \textbf{Strict evaluation} (Rel+) further requires the model to predict 
the entity types on the basis of the requirement of the boundary prediction. Moreover, following \citet{UniRE}, we regard each symmetric relational instance as two directed relational instances.

 
 


\subsubsection{Implementation Details}

\begin{table}[!t]
\small
\resizebox{0.48\textwidth}{!}{
    \centering
    \begin{tabular}{l|rrc}
    \toprule
    \textbf{Dataset} &  \makecell[c]{\textbf{\#Sents}}  &
     \makecell[c]{\textbf{\#Ents (\#Types)}}
     & \textbf{\#Rels (\#Types)}  \\
    \midrule
     CoNLL03 & 22.1k   &35.1k \,\,\,(4) & -  \\
     OntoNotes 5.0& 103.8k   &161.8k (18) & -  \\
     Few-NERD & 188.2k   &491.7k (66) & - \\
     ACE05 & 14.5k   & 38.3k  \,\,\,(7)& 7.1k (6) \\
     ACE04 & 8.7k   & 22.7k \,\,\,(7) & 4.1k (6)  \\
     SciERC & 2.7k   &8.1k \,\,\,(6) & 4.6k (7) \\
    \bottomrule
    \end{tabular}
    }
 \caption{The statistics of the adopted datasets.}
 
 
 \label{tab:statistics}

\end{table}

We adopt \emph{bert-base-uncased}~\cite{BERT}  and \emph{albert-xxlarge-v1}~\cite{albert}  encoders for ACE04 and ACE05. For SciERC, we  use the in-domain \emph{scibert-scivocab-uncased}~\cite{scibert}  encoder. For flat NER, we adopt \emph{roberta-large} encoder.  
We also leverage the cross-sentence information~\cite{DyGIE, crosssentencener}, which extends each sentence by its context and ensures that the original sentence is located in the middle of the expanded sentence as much as possible. 
As discussed in Section~\ref{sec:compare_ner}, for the packing scheme on NER,  we set the group size to 256 to improve efficiency. 
We run all experiments with 5 different seeds and report the average score. See the appendix for the standard deviations and the detailed training configuration.






\subsection{Named Entity Recognition}

\subsubsection{Baselines}
Our packing scheme allows the model to apply the levitated markers to process massive span pairs and to our best knowledge, we are the first to apply the levitated markers on the NER task. We compare our neighborhood-oriented packing scheme with the \textbf{Random Packing}, which randomly packs the candidate spans into groups. We adopt two common NER models: (1) \textbf{SeqTagger}~\cite{BERT} regards NER as a sequence tagging task and applies a token-level classifier to distinguish the IOB2 tags for each word~\cite{BIO}. 
(2) \textbf{T-Concat}~\cite{Generalspan, PURE}  assigns an entity type or a non-entity type to each span based on its T-Concat span representation. 
Note that solid markers cannot deal with the overlapping spans simultaneously, thus it is too inefficient to apply solid markers independently on the NER task.

\subsubsection{Results}

%

\begin{table}[!t]
 \resizebox{0.48\textwidth}{!}{

    \centering
    \begin{tabular}{l|c|c|c}
        \toprule
        \textbf{Model} & {\textbf{CoNLL03}} & {\textbf{OntoN5}} & {\textbf{F-NERD}}\\
        \midrule
        \citet{BiLSTM-CRF} & 91.0&86.3 & - \\
        \citet{BERT}& 92.8& 89.2 & {68.9} \\ 
\citet{BERTMRC} & 93.0  &91.1 &-\\
\citet{DependencyParsingNER} & 93.5& 91.3 &-\\
\citet{BARTNER} & 93.2 & 90.4 & - \\
        
      SeqTagger\,$_{\small \text{(Our impl.)}}$ &  93.6 & 91.2  &  69.0 \\
        T-Concat\,$_{\small \text{(Our impl.)}}$  & 93.0 &  {91.7} & 70.6 \\
        \midrule
        Random Packing & {93.9} &  {91.8} &  {61.5} \\
         \Ourmodel \,$_{\small \text{(Our model)}}$  &  \textbf{94.0}&  \textbf{91.9} &  \textbf{70.9} \\
        \bottomrule
    \end{tabular}
    }
 \caption{Micro F1 on the test set for the flat NER. OntoN5: OntoNotes 5.0; F-NERD: Few-NERD. 
 }  

 \label{tab:flatner}

\end{table}

We show  the flat NER results in the Table~\ref{tab:flatner} and the nested NER results in the Ent column of  Table~\ref{tab:re}, where PURE~\cite{PURE}  applies the T-Concat feature on its NER module. As follow, some observations are summarized from the experimental results: 
(1) The model with our neighborhood-oriented packing strategy outperforms the model with random packing strategy on all three flat NER datsets, especially obtaining a 9.4\% improvement on Few-NERD.  Few-NERD contains longer sentences and thus includes  325 candidate spans on average, while CoNLL03 and OntoNotes 5.0 only contain 90 and 174 respectively. It shows that the neighborhood-oriented packing strategy can well handle the dataset with longer sentences and more groups of markers, to better model the interrelation among neighbor spans. 
(2) With the same large pre-trained encoder,  \Ourmodel achieves an absolute F1 improvement of +0.1\%-1.1\%  over T-Concat on all six NER benchmarks, which  shows the advantage of levitated markers in aggregating span-wise representation for the entity type prediction; 
(3) \Ourmodel outperforms SeqTagger by  an absolute F1 of +0.4\%, +0.7\%, +1.9\% in CoNLL03, OntoNote 5.0 and  Few-NERD respectively, where  CoNLL03, OntoNote 5.0 and Few-NERD contain 4, 18 and 66 entity types respectively. Such improvements prove the effectiveness of \Ourmodel in handling diverse interrelation between entities of diverse types. 



\subsection{Relation Extraction}

\begin{table*}[!t]
 \resizebox{0.995\textwidth}{!}{
    \centering
    \begin{tabular}{l|c|c|ccc|ccc|ccc}
        \toprule
        \multirow{2}{*}{\textbf{Model}} & \multirow{2}{*}{\textbf{Encoder}} & \textbf{Rep}& \multicolumn{3}{c|}{\textbf{ACE05}} & \multicolumn{3}{c|}{\textbf{ACE04}} & \multicolumn{3}{c}{\textbf{SciERC}} \\
        && \textbf{Type}& \textbf{Ent} & \textbf{Rel} & \textbf{Rel+} 
        &  \textbf{Ent} & \textbf{Rel} & \textbf{Rel+}  
        &  \textbf{Ent} & \textbf{Rel} & \textbf{Rel+} \\
       \midrule
\citet{qili} & - & -& 80.8& 52.1& 49.5& 79.7& 48.3& 45.3& -& -& -\\
SPtree~\cite{SPtree} & LSTM &T& 83.4 & - &55.6 &81.8 &- &48.4 &- &- &-\\
DYGIE~\cite{DyGIE}$^{\Diamond}$ &  ELMo &T& 88.4 &63.2& - &87.4 &59.7& - &65.2& 41.6& - \\
\midrule
Multi-turn QA~\cite{multiturnQA} &  \multirow{2}{*}{BERT\bl}& -&84.8& -& 60.2& 83.6& -& 49.4& -& -& - \\
OneIE~\cite{OneIE} & & T &  88.8& 67.5&-& -& -& -& -& -&- \\
       \midrule

DYGIE++~\cite{DyGIEpp}$^{\Diamond}$ &  \multirow{6}{*}{\makecell[c]{BERT\bb /  \\ SciBERT}}  & T
&88.6 & 63.4&-& -& -& -& -& -&- \\
TriMF~\cite{TriMF}$^{\Diamond}$ &  & T& 87.6 & 66.5 &62.8 & -& -& - & \textbf{70.2} & {52.4} &  -   \\
UniRE~\cite{UniRE}$^{\Diamond}$ & & T& 88.8& - &  64.3 &  87.7 & - &60.0 & 68.4 & -& 36.9   \\
PURE-F~\cite{PURE}$^{\Diamond}$ & & S& \textbf{90.1}&67.7&  64.8 &  \textbf{89.2} &63.9 &60.1 & 68.9 & 50.1& 36.8  \\
PURE-A~\cite{PURE}$^{\Diamond}$ & & L& -&66.5&  - &  - &- &- &-& 48.1& - \\
\Ourmodel (Our Model)$^{\Diamond}$ &  & S\&L & 89.8  & \textbf{69.0}  & \textbf{66.5}  & {88.8}  & \textbf{66.7} &  \textbf{62.6} & 69.9  & \textbf{53.2}   & \textbf{41.6}  \\
\midrule
TableSeq~\cite{twobetterthanone} & \multirow{4}{*}{ALB\xxl} & T & 89.5 & 67.6 &64.3& 88.6 &63.3 &59.6 & - & - & - \\
UniRE~\cite{UniRE}$^{\Diamond}$  & & T& 90.2 & - & 66.0 & 89.5& -& 63.0 & - & -& -  \\
PURE-F~\cite{PURE}$^{\Diamond}$  &  & S& 90.9 & 69.4 & 67.0 & {90.3}& 66.1& 62.2 & - & -&  -  \\
\Ourmodel (Our Model)$^{\Diamond}$ &  & S\&L& \textbf{91.1} & \textbf{73.0} & \textbf{71.1} & \textbf{90.4} &  \textbf{69.7} & \textbf{66.5}   & - & -&  -  \\

    \bottomrule
    \end{tabular}
}
 \caption{Overall entity and relation F1 scores on the test sets of ACE04, ACE05 and SciERC. The encoders used in different models: BERT\bb=BERT\base, BERT\bl = BERT\blarge, ALB\xxl = ALBERT\xxlarge.   Specially, TriMF, UniRE, PURE and \Ourmodel apply BERT\base on ACE04/05 and apply the SciBERT on SciERC. $^{\Diamond}$ denotes that the model leverages the cross-sentence information.  Representation Type: T--\emph{T-Concat}; S--\emph{Solid Marker}; \emph{L}--\emph{Levitated Marker}. Model name abbreviation: PURE-F: PURE (Full); PURE-A: PURE (Approx.).
 }
 \label{tab:re}
\end{table*}
\subsubsection{Baselines} 
For the end-to-end RE, we compare our model,  \Ourmodel,  with a series of state-of-the-art models. Here, we introduce two of the most representative works  with  T-Concat and Solid Markers span representation: (1) \textbf{DyGIE++}~\cite{DyGIEpp} first acquires the T-Concat span representation, and then iteratively propagates coreference and relation type confidences  through a span  graph to refine the representation; 
(2)  \textbf{PURE}~\cite{PURE} adopts independent NER and RE models, where the RE model processes each possible entity pair in one pass. In their work, PURE (Full)  adopts two pairs of solid markers to emphasize a span pair and the PURE (Approx) employs two pairs of levitated markers to underline the span pair. 


\subsubsection{Results}

As shown in Table~\ref{tab:re},  with the same BERT\base encoder, our approach outperforms previous methods by strict F1 of +1.7\% on ACE05 and +2.5\% on ACE04. With the SciBERT encoder, our  approach also achieves the best performance on SciERC. Using a larger encoder,  ALBERT\xxlarge, both of our NER and RE models are further improved. Compared to the previous state-of-the-art model,  PURE (Full), our model gains a substantially +4.1\% and  +4.3\%  strict relation F1 improvement on ACE05 and ACE04 respectively. 
Such improvements over PURE indicate the effectiveness of modeling the interrelation between the same-subject or the same-object entity pairs in the training process.

\subsection{Inference Speed}
\label{sec:compare_speed}
In this section, we compare the models' inference speed on an  A100 GPU with a batch size of 32. We use the \texttt{BASE} size encoder for ACE05 and SciERC in the experiments and  the \texttt{LARGE} size encoder for flat NER models.

\subsubsection{Speed of Span Model}
\label{sec:compare_ner}

We evaluate the inference speed of \Ourmodel with different group size $K$ on CoNLL03 and Few-NERD.  We also evaluate a cascade \textbf{Two-stage} model, which  uses a fast \texttt{BASE}-size T-Concat model to filter candidate spans for our model.
As shown in  Table~\ref{tab:nerspeed},  \Ourmodel achieves a 0.4 F1 improvement on CoNLL03 but sacrifices 60\% speed compared to the SeqTagger model. 
And we observe that our proposed Two-stage model achieves similar performance to \Ourmodel with  3.1x speedup on Few-NERD, which shows it is more efficient to use \Ourmodel as a post-processing module to elaborate the coarse prediction from a simple model.  
In addition, when the group size grows to 512, \Ourmodel slows down due to the increased complexity of the Transformer. Hence, we choose a group size of 256 in practice. 

\begin{table}[!t]
\small
\centering
\begin{tabular}{l|c|cc|cc}
\toprule
\multirow{3}{*}{\textbf{Model}} & \multirow{3}{*}{{$\mathbf{K}$}}  & \multicolumn{2}{c|}{\textbf{CoNLL03}} & \multicolumn{2}{c}{\textbf{Few-NERD}} \\
&& Ent & Speed & Ent & Speed \\
&& (F1) & (sent/s) & (F1) & (sent/s) \\ 
\midrule
SeqTagger & - & \textbf{93.6}& \textbf{138.7} & 69.0 & \textbf{142.0} \\
T-Concat & - &93.0 & 137.2 & \textbf{70.6} & 126.8 \\   
\midrule
\multirow{3}{*}{{\Ourmodel}} 
& 128 & \textbf{94.0}& \textbf{54.8} & \textbf{70.9}  &  23.8 \\     
& 256 &-& 39.6 & - & \textbf{25.8} \\   
& 512 &-& 22.9  & - & {18.3}\\
\midrule
\multirow{2}{*}{{Two-stage}} 
& \,\,16 & 93.7 & \textbf{87.1} & 70.8  & \textbf{80.6} \\ 
& \,\,32 & \textbf{94.0} & 83.3 &   \textbf{70.9} & 79.8\\ 
\bottomrule
\end{tabular}
 \caption{Micro F1 and efficiency on NER benchmarks with respect to the model and  different packing group size $K$.  We adopt a maximum span length of 8 for CoNLL03 and 16 for Few-NERD. 
 }

 \label{tab:nerspeed}

\end{table}

\begin{table}[!t]
\small
\centering
\begin{tabular}{l|cc|cc}
\toprule
\multirow{3}{*}{\textbf{Model}} & \multicolumn{2}{c|}{\textbf{ACE05}} & \multicolumn{2}{c}{\textbf{SciERC}} \\
& Rel & Speed & Rel & Speed \\
& (F1) & (sent/s) & (F1) & (sent/s) \\
\midrule
PURE (Full) & 67.7 &  76.5 & 50.1 &  88.3 \\
PURE (Approx.)& 66.5 & \textbf{593.7} &48.8 & \textbf{424.2}  \\
\Ourmodel &  \textbf{69.3} & 211.7  & \textbf{52.8}  & 190.9 \\
\bottomrule
\end{tabular}
 \caption{Comparison of our RE model and  PURE in relation F1 (boundaries) and speed. We report the result with \texttt{BASE} encoders. All models adopt the same entity input from the entity model of PURE. 
 }

 \label{tab:respeed}

\end{table}

\subsubsection{Speed of Span Pair Model}
\label{sec:compare_pure}

We apply the subject-oriented and the object-oriented  packing strategies on levitated markers for RE. Here, we compare our model with the other two marker-based models. Firstly, \textbf{PURE (Full)}~\cite{PURE} applies solid markers to  process each entity pair independently. Secondly,  \textbf{PURE (Approx.)}  packs the levitated  markers of all entity pairs into an instance for  batch computation. 
Since the performance and the running time of the above methods rely on the quality and the number of predicted entities, for a fair comparison, we adopt the same entity input from the entity model of PURE on all the RE models. 
Table~\ref{tab:respeed} shows the relation F1 scores and the inference speed of the above three methods. On both datasets,  our RE model, \Ourmodel, achieves the best performance and PURE (Approx.)  has highest efficiency in the inference process.  Compared to the PURE (Full), our model obtains a 2.2x-2.8x speedup and better performance on ACE05 and SciERC. 
Compared to PURE (Approx.), our model achieves a  2.8\%-4.0\%  relation F1 (boundaries) improvement on ACE05 and SciERC,  which again demonstrates the effectiveness of our fusion markers and packing strategy. 
Overall, our model, with a novel subject-oriented packing strategy for markers, has been proven effective in practice, with satisfactory accuracy and affordable cost.

\subsection{Case Study}

We show several cases to compare our span model with T-Concat and to compare our span pair model with PURE (Full). As shown in Table~\ref{tab:casestudy}, our span model could collect contextual information, such as \emph{Taiwan} and \emph{mainland}, for underlined span, \emph{Cross Strait}, assisting in predicting its type as organization rather than work of art.  
Our span model learns to integrally consider the interrelation between the same-object relational facts in  training phase, so as to successfully obtain the fact that both \emph{Liana} and \emph{her parents} are located in \emph{Manhattan}.

\begin{table}[!t]
\small

\centering
\begin{tabular}{p{0.94\columnwidth}}
\toprule
\textbf{Named Entity Recognition} \\
\midrule
\textbf{Text}: This is the Cross Strait program on CCTV International Channel. ... Candidates for the giant pandas to be presented to Taiwan as gifts from the mainland may increase.  ...  \\
\textbf{T-Concat}: \textcolor{blue}{ (Cross Strait, WORK OF ART)}, (CCTV International Channel, ORG), (Taiwan, GPE)\\
\textbf{Our}: \textcolor{red}{(Cross Strait, ORG)}, (CCTV International Channel, ORG), (Taiwan, GPE)\\
\midrule
\textbf{Relation Extraction} \\
\midrule
\textbf{Text}: \emph{Liana} drove 10 hours from \emph{Pennsylvania} to attend the rally in \emph{Manhattan} with \emph{her parents} \\

\textbf{PURE}: (Liana, located in, Manhattan) \\
\textbf{Our}: \,\,\,\,{(Liana, located in, Manhattan),  \textcolor{red}{(her parents, located in, Manhattan)}} \\
\bottomrule
\end{tabular}
 \caption{Case study of our NER and RE model.}
 \label{tab:casestudy}

\end{table}

\subsection{Ablation Study}

In this section, we conduct ablation studies  to investigate the contribution of different components to our RE model, where we apply  \texttt{BASE} size encoder in the experiments.

\begin{table}[!t]
\small

\centering
\begin{tabular}{l|cc|cc}
\toprule
\multirow{3}{*}{\textbf{Model}} & \multicolumn{2}{c|}{\textbf{ACE05}} & \multicolumn{2}{c}{\textbf{SciERC}} \\
 & gold & e2e  & gold & e2e \\
\midrule
\Ourmodel &  \textbf{74.0}  & \textbf{69.0} & \textbf{72.5}  & \textbf{53.2} \\
\,\,w/o. solid marker  & 72.0 & 67.3 & 68.7 &  50.6  \\
\,\,w/o. inverse relation  & 72.9  &  68.1  &  71.6 & 52.7    \\
\,\,w/o. entity type loss  &  73.4 & 68.4  & 72.3  &  \textbf{53.2} \\
\,\,w.\,\, type marker &\textbf{74.0}& 68.3& 72.1 & 53.0 \\
\bottomrule
\end{tabular}
 \caption{The relation F1 (boundaries) on the test set of ACE05 and SciERC with different input features for the ablation study. gold: use the gold entities; e2e: use the entities predicted by our entity model. w/o.: without. w.: with.}
 \label{tab:ablation_re}
\end{table}

\paragraph{Two pairs of Levitated Markers}
We evaluate the \emph{w/o solid marker} baseline, which applies two pairs of levitated markers on the subject and object respectively and packs all the span pairs into an instance. 
As shown in Table~\ref{tab:ablation_re}, compared to \Ourmodel, the model without solid markers drops a huge 2.0\%-3.8\% F1 on ACE05 and SciERC when the golden entities are given. The result demonstrates that it is sub-optimal to continue to apply directional attention to bind two pairs of levitated markers, since a pair of levitated marker is already tied by the   directional attention.

\paragraph{Inverse Relation} We establish an inverse relation for each asymmetric relation for a bidirectional prediction.  We evaluate the model without inverse relation, which replaces the constructed inverse relation with a non-relation type and adopts a unidirectional prediction.
As shown in  Table~\ref{tab:ablation_re}, the model without inverse relation drops 0.9\%-1.1\% F1  on both datasets with the gold entities given,  indicating the significance of modeling the information from the object entity to the subject entity in our asymmetric framework.


\paragraph{Entity Type} We add an auxiliary entity type loss to RE model to introduce the entity type information. As shown in  Table~\ref{tab:ablation_re}, 
when the gold entities are given, the model without entity type loss drops 0.4\%-0.7\% F1 on both datasets, which shows the importance of entity type information in RE.  
Moreover, we try to apply the type markers~\cite{PURE}, such as \emph{[Subject:PER]} and \emph{[Object:GPE]}, to inject entity type information predicted by the NER model into the RE model. We find the RE model with type marker performs slightly worse than the model with entity type loss in the end-to-end setting. It shows that the entity type prediction error from the NER model may be propagated to the RE model if we adopt the type markers as input features. 
Finally, we discuss when to use  the entity type prediction from  the RE model to refine the NER prediction in the Appendix and we finally  refine entity type for ACE04 and ACE05 except SciERC according to their dataset statistic. 




\section{Conclusion}
In this work, we present a novel packed levitated markers, with a neighborhood-oriented packing strategy and a subject-oriented packing strategy, to obtain the span (pair) representation. Considering the interrelation between spans  and span pairs,  our model achieves the state-of-the-art F1 scores and a promising efficiency on both NER and RE tasks across six standard benchmarks. In future, we will further investigate how to generalize the marker-based span representation to more NLP tasks.

\section*{Acknowledgement}
This work is supported by the National Key R\&D Program of China (No.\,2020AAA0106502), Institute Guo Qiang at Tsinghua University, and International Innovation Center of Tsinghua University, Shanghai, China.  We thank Chaojun Xiao and other members of THUNLP for
their helpful discussion and feedback. Deming Ye conducted the experiments. Deming Ye, Yankai Lin, Xiaojun Xie and Peng Li wrote the paper. Maosong Sun provided valuable advices to the research. 

\bibliography{anthology,custom}
\bibliographystyle{acl_natbib}

\appendix
\newpage
\section{Training Configuration}




We train all the models with Adam optimizer~\cite{Adam} and 10\% warming-up steps. And we adopt a learning rate of 2e-5 for  models with \texttt{BASE} size and a learning rate of 1e-5 for models with \texttt{LARGE} or \texttt{XXLARGE} size.  We run all experiments with 5 different seeds (42, 43, 44, 45, 46).   
For the bidirectional prediction on RE, we set the forward and inverse relation of symmetric labels to be consistent.  The symmetric labels include  the \emph{PER-SOC}   in the ACE04/ACE05  and  the \emph{Compare} and \emph{Conjunction}  in the SciERC. 

For NER model, we set the maximum length of expanded sentence $C$ as 512.  For RE model, we set $C$ as 256 for ACE05 and SciERC and set  $C$ as 384 for ACE04. To enumerate possible spans, we set the maximum span length $L$  as 16 for OntoNote 5.0 and Few-NERD and set 8 for the other datasets. 
For the NER on CoNLL03 and the RE on ACE05, we search the batch size in [4,8,16,32] and observe the model with a batch size of 8  achieves a sightly better performance. Hence, we choose a batch size of 8 for all the datasets. We search the number of epochs in [3,5,8,10,15,50] for all the datasets and finally choose  8 for CoNLL03, 4  for OntoNote 5.0, 3 for Few-NERD, 10 for ACE05-NER, 15 for ACE04-NER, 50 for SciERC-NER and 10 for all the RE models.


\section{Prompt Initialization for NER}

Inspired by the success of {prompt} tuning~\cite{GPT3,PET},  we use the embedding of meaningful words instead of randomness to initialize the embedding of markers for the NER models. To be specific, we initialize a pair of markers for the span with the words \texttt{[MASK]} and \emph{entity}.  As shown in  Table~\ref{tab:ablation_ner},  using meaningful words as prompt to initialize the markers can bring a slight improvement to all six NER benchmarks.

\begin{figure}[!t]
    \centering
    \includegraphics[width=\linewidth]{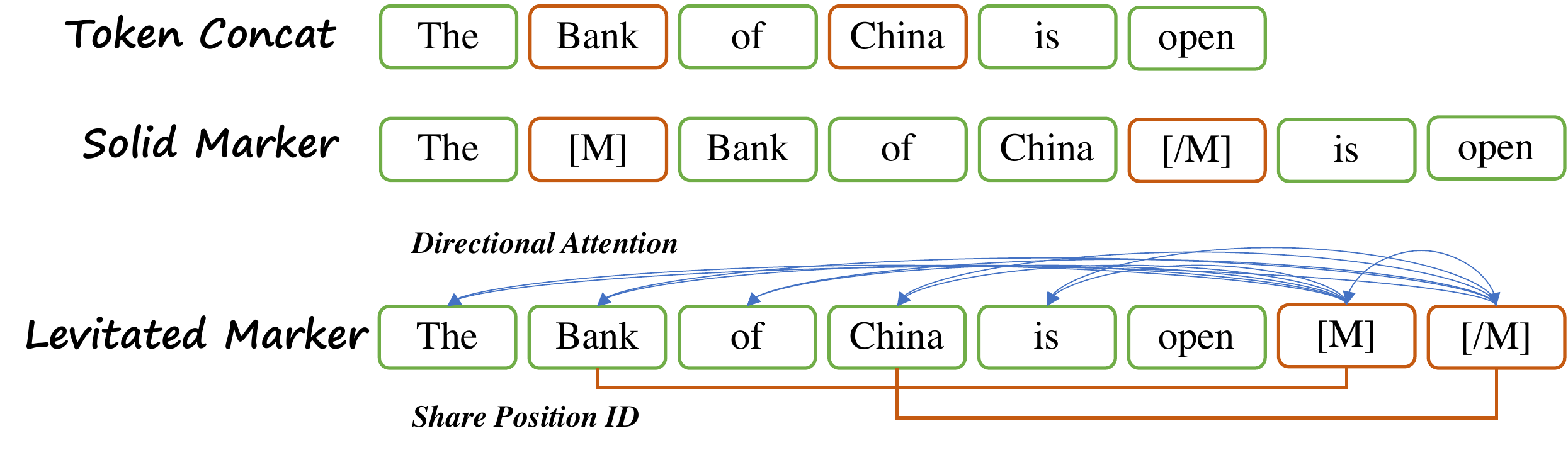}
    \caption{
     Span representation  of T-Concat, Solid Marker and Levitated Marker. We highlight the attention direction for the levitated marker and omit the bidirectional attention line between other token pairs.  Token Concat conveys the representation of edged token, [Bank] and [China], through the classifier, while Solid Marker and Levitated Marker employ the features of two markers, [M] and [/M]  for classification.
    }
    \label{fig:spanrep}
\end{figure}

\begin{table}[!t]
\resizebox{0.48\textwidth}{!}{
    \centering
    \begin{tabular}{l|ccc}
        \toprule
        \textbf{Init Strategy} & \textbf{CoNLL03} &  \textbf{OntoNote 5.0} & \textbf{Few-NRED}\\
       \midrule
Random & {93.9}$_{\pm {0.1}}$  & {{91.8}}$_{\pm {0.0}}$ & {70.8}$_{\pm {0.1}}$\\
Prompt & \textbf{94.0}$_{\pm {0.1}}$ &  \textbf{91.9}$_{\pm {0.2}}$ & \textbf{70.9}$_{\pm {0.1}}$ \\
\midrule
        \textbf{Init Strategy} & \textbf{ACE05} &  \textbf{ACE04} & \textbf{SciERC}\\
\midrule
Random &   {{91.0}}$_{\pm {0.3}}$ &{90.3}$_{\pm {0.5}}$  & {69.4}$_{\pm {0.5}}$\\
Prompt & \textbf{91.1}$_{\pm {0.3}}$ &   \textbf{90.4}$_{\pm {0.4}}$  &  \textbf{69.9}$_{\pm {0.7}}$  \\

    \bottomrule
    \end{tabular}
    }
 \caption{The entity F1 on test set of NER datasets when the \Ourmodel is initialized with prompt or when it is initialized randomly. We use the RoBERTa\blarge for flat NER datasets and ALBERT\xxlarge for the nested NER datasets. }
 
 \label{tab:ablation_ner}
\end{table}

\section{Refine Entity Type}

We  attempt to apply the entity type prediction from  the RE model to refine the entity type prediction from  the NER model. As shown in  Table~\ref{tab:ablation_re_refile}, using the entity type predicted by the RE model brings +0.5\% and -0.9\% strict relation F1 on ACE05 and SciERC respectively. We observe that the most frequent entity type pair for each relation occupies 48.5\%  for ACE05, 52.0\% for ACE04 and 19.1\% for SciERC, which shows that the relation is more relevant to the entity type in ACE05 than that in SciERC. Hence,  we only use the RE model to refine the NER results for ACE04 and ACE05.

\begin{table}[!t]
\small

\centering
\begin{tabular}{l|cc|cc}

\toprule
\multirow{2}{*}{\textbf{Entity Type Source}} & \multicolumn{2}{c|}{\textbf{ACE05}} & \multicolumn{2}{c}{\textbf{SciERC}} \\
 & Ent & Rel+  & Ent & Rel+ \\
\midrule
NER Model  & \textbf{89.8}  &  66.0 &   \textbf{69.9}  & \textbf{41.6}   \\
+RE Model Refine &  \textbf{89.8} & \textbf{66.5} & {69.5} & 40.7 \\
\bottomrule
\end{tabular}
 \caption{The entity F1 and the strict relation F1 on the test set of ACE05 and SciERC   when the RE model is used to refine the NER prediction or when it is not used. }
 \label{tab:ablation_re_refile}

\end{table}

\begin{table}[!t]
\small
\centering
\begin{tabular}{l|cc|cc}
\toprule
\multirow{3}{*}{\textbf{Model}} & \multicolumn{2}{c|}{\textbf{ACE05}} & \multicolumn{2}{c}{\textbf{SciERC}} \\
& Rel & Speed & Rel & Speed \\
& (F1) & (sent/s) & (F1) & (sent/s) \\
\midrule
PURE  & 67.7 &  76.5 & 50.1 &  88.3 \\
PURE w. InvRel.& 68.4 & 76.2 & 52.5 & 87.9  \\
\midrule
\Ourmodel &  \textbf{69.3} & \textbf{211.7}  & \textbf{52.8}  & \textbf{190.9} \\
\bottomrule
\end{tabular}
 \caption{Comparison of our RE model and  PURE in relation F1 (boundaries) and speed. All models adopt the same entity input from the entity model of PURE. 
 }

 \label{tab:respeed2}

\end{table}

\section{Inverse Relation on Baseline}
We apply the inverse relation and bidirectional prediction on the baseline PURE (Full)~\cite{PURE} to obtain the PURE w. InvRel. model. As shown in Table~\ref{tab:respeed2}, except for our asymmetric framework, the bidirectional prediction can also improve the symmetrical baseline PURE by 0.7\%-2.4\% relation F1 on ACE05 and SciERC. 


\section{Detailed NER Results}
We illustrate the span representation adopted by the NER models in Figure~\ref{fig:spanrep}.  And we show the average scores of the baselines and \Ourmodel  on flat NER with standard deviations in Table~\ref{tab:ner2}.

\begin{table}[!t]
\resizebox{0.48\textwidth}{!}{

    \centering
    \begin{tabular}{l|c|c|c}
        \toprule
        \textbf{Model} & {\textbf{CoNLL03}} & {\textbf{OntoNotes\,5}} & {\textbf{Few-NERD}}\\
        \midrule
      SeqTagger &  93.6$_{\pm {0.1}}$ & 91.2$_{\pm {0.2}}$  &  69.0$_{\pm {0.1}}$ \\
        Token Concat & 93.0$_{\pm {0.2}}$ &  {91.7}$_{\pm {0.1}}$ & 70.6$_{\pm {0.1}}$ \\
        \midrule
        Random Packing & {93.9}$_{\pm {0.2}}$ &  {91.7}$_{\pm {0.2}}$ &  {61.5}$_{\pm {0.1}}$ \\
         \Ourmodel &  \textbf{94.0}$_{\pm {0.1}}$ &  \textbf{91.9}$_{\pm {0.1}}$ &  \textbf{70.9}$_{\pm {0.1}}$ \\
        \bottomrule
    \end{tabular}
    }
 \caption{Overall entity  F1 scores of the baselines  and \Ourmodel on the test set of CoNLL03, OntoNotes 5.0 and Few-NERD. We report average scores across five random seeds, with standard deviations as subscripts.
 }

 \label{tab:ner2}
\end{table}

\begin{table}[!t]
\small

    \centering
    \begin{tabular}{l|c|ccc}
        \toprule
        \textbf{Dataset} & \textbf{Encoder} 
        &  \textbf{Ent} & \textbf{Rel} & \textbf{Rel+} \\
       \midrule
      \multirow{2}{*}{\textbf{ACE05}} & BERT\bb & 89.8$_{\pm 0.2}$  & {69.0}$_{\pm 0.5}$   & {66.5}$_{\pm 0.4}$  \\
      & ALB\xxl & \textbf{91.1}$_{\pm 0.2}$ & \textbf{73.0}$_{\pm 0.9}$ & \textbf{71.1}$_{\pm 0.6}$ \\
     \midrule
      \multirow{2}{*}{\textbf{ACE04}} & BERT\bb & {88.8}$_{\pm 0.8}$  & {66.7}$_{\pm 1.1}$ &  {62.6}$_{\pm 1.3}$   \\
      & ALB\xxl & \textbf{90.4}$_{\pm 0.5}$ & \textbf{69.7}$_{\pm 1.9}$ & \textbf{66.5}$_{\pm 2.2}$ \\
       \midrule
      \textbf{SciERC} & SciBERT &  \textbf{69.9}$_{\pm 0.7}$  & \textbf{53.2}$_{\pm 0.9}$   & \textbf{41.6}$_{\pm 0.8}$  \\

    \bottomrule
    \end{tabular}
 \caption{Overall entity and relation F1 scores of \Ourmodel on the test set of ACE04, ACE05 and SciERC. BERT\bb denotes BERT\base; ALB denotes ALBERT\xxlarge; We report average scores across five random seeds with standard deviations as subscripts.}
 
 \label{tab:re2}
\end{table}

\section{Detailed RE Results}
We show the average scores of \Ourmodel on RE with standard deviations in Table~\ref{tab:re2}.

\end{document}